\documentclass[11pt]{article}

\usepackage[font=footnotesize]{caption}

\usepackage{naaclhlt2016}

\makeatletter
\newcommand{\@BIBLABEL}{\@emptybiblabel}
\newcommand{\@emptybiblabel}[1]{}
\makeatother
\usepackage{hyperref}

\usepackage{times}
\usepackage{latexsym}
\usepackage{amsmath}
\usepackage{multirow}
\usepackage{amsfonts}

\usepackage{textcomp}
\usepackage{color}
\usepackage{fancyvrb}
\usepackage{graphicx}
\usepackage{tabularx}
\usepackage{umoline}

\newcommand{\mbf}[1]{\mathbf{#1}}

\definecolor{green}{rgb}{0.1,0.5,0.1}
\definecolor{red}{rgb}{1.0,0.2,0.2}

\newcommand{\remove}[1]{}
\newcommand{\add}[1]{{#1}}
\newcommand{\figpath}{}

\naaclfinalcopy 
\def\naaclpaperid{***} 

\title{Semi-supervised Question Retrieval with Gated Convolutions}

\author{
    Tao Lei$\quad$Hrishikesh Joshi$\quad$Regina Barzilay$\quad$Tommi Jaakkola \\
    \vspace{-0.05in}MIT{~} CSAIL\\
    \vspace{0.12in}{\small\tt \{taolei, hjoshi, regina, tommi\}@csail.mit.edu}\\ 
    \vspace{-0.02in}$\quad$\textbf{Katerina Tymoshenko $\ \ \ \quad\quad${~~~}Alessandro Moschitti $\ \ $Llu{\'\i}s M{\`a}rquez}\\
    {~~~}$\quad\ \ \ $University of Trento $\qquad\ \ \ \ \ $ Qatar Computing Research Institute, HBKU\\
    {\small\tt tymoshenko@disi.unitn.it}  $\quad\ \ \ $ {\small\tt \{amoschitti, lmarquez\}@qf.org.qa} 
}
\date{}

\begin{document}

\maketitle

\begin{abstract}
Question answering forums are rapidly growing in size with no effective automated ability to refer to and reuse answers already available for previous posted questions. In this paper, we develop a methodology for finding semantically related questions. The task is difficult since 1) key pieces of information are often buried in extraneous details in the question body and 2) available annotations on similar questions are scarce and fragmented. We design a recurrent and
convolutional model (gated convolution) to effectively map questions to their semantic representations. The models are pre-trained within an encoder-decoder framework (from body to title) on the basis of the entire raw corpus, and fine-tuned discriminatively from limited annotations.  Our evaluation demonstrates that our model yields substantial gains over a standard IR baseline and various neural network architectures (including CNNs, LSTMs and GRUs).\footnote{Our code and data are available at \url{https://github.com/taolei87/rcnn}}
\end{abstract}

\section{Introduction}

Question answering (QA) forums such as Stack Exchange\footnote{\url{http://stackexchange.com/}} are rapidly expanding and already contain millions of questions. The expanding scope and coverage of these forums often leads to many duplicate and interrelated questions, resulting in the same questions being answered multiple times. By identifying similar questions, we can potentially reuse existing answers, reducing response times and unnecessary repeated work. Unfortunately in most forums, the process of identifying and referring to existing similar questions is done manually by forum participants with limited, scattered success. 

The task of automatically retrieving similar questions to a given user's question has recently attracted significant attention 
and has become a testbed for various representation learning approaches~\cite{zhou-EtAl:2015,dossantos-EtAl:2015}. However, the task has proven to be quite challenging -- for instance, \newcite{dossantos-EtAl:2015} report a 22.3\% classification accuracy, yielding a 4 percent gain over a simple word matching baseline. 

\begin{figure}[t]
\vspace{-0.03in}
\centering
\footnotesize
\begin{tabular}{@{}c@{}|c|@{}c@{}}
\cline{2-2}
\vspace{-0.06in}& & \\
&
\begin{minipage}{0.43\textwidth}
\textbf{Title:} How can I boot Ubuntu from a USB?

\textbf{Body:} I bought a Compaq pc with Windows 8 a few months ago and now I want to install Ubuntu but still keep Windows 8. I tried Webi but when my pc restarts it read ERROR 0x000007b. I know that Windows 8 has a thing about not letting you have Ubuntu but I still want to have both OS without actually losing all my data ...
\end{minipage} & \\
\vspace{-0.06in} & & \\
\cline{2-2}
\vspace{-0.06in} & & \\
&
\begin{minipage}{0.43\textwidth}
\textbf{Title:} When I want to install Ubuntu on my laptop I'll have to erase all my data. ``Alonge side windows'' doesnt appear

\textbf{Body:} I want to install Ubuntu from a Usb drive. It says I have to erase all my data but I want to install it along side Windows 8. The ``Install alongside windows'' option doesn't appear. What appear is, ...
\end{minipage} & \\
& & \\
\cline{2-2}
\end{tabular}
\caption{A pair of similar questions.}
\label{fig:example}
\vspace{-0.2in}
\end{figure}

Several factors make the problem difficult. First, submitted questions are often long and contain extraneous information irrelevant to the main question being asked. For instance, the first question in Figure~\ref{fig:example} pertains to booting Ubuntu using a USB stick. A large portion of the body contains tangential details that are idiosyncratic to this user, such as references to \textit{Compaq pc}, \textit{Webi} and the error message. Not surprisingly, these features are not repeated in the second question in Figure~\ref{fig:example} about a closely related topic. The extraneous detail can easily confuse simple word-matching algorithms. Indeed, for this reason, some existing methods for question retrieval 
restrict attention to the question title only. While titles (when available) can succinctly summarize the intent, they also sometimes lack crucial detail available in the question body. For example, the title of the second question does not refer to installation from a USB drive. The second challenge arises from the noisy annotations. Indeed, the pairs of questions marked as similar by forum participants are largely incomplete. Our manual inspection of a sample set of questions from AskUbuntu\footnote{\url{http://askubuntu.com/}} shows that only 5\% of similar pairs have been annotated by the users, with a precision of around 79\%.

In this paper, we design a neural network model and an associated training paradigm to address these challenges. On a high level, our model is used as an encoder to map the title, body, or the combination to a vector representation. The resulting ``question vector'' representation is then compared to other questions via cosine similarity. We introduce several departures from typical architectures on a finer level. In particular, we incorporate adaptive gating in
non-consecutive CNNs~\cite{Lei15} in order to focus temporal averaging in these models on key pieces of the questions. Gating plays a similar role in LSTMs~\cite{hochreiter1997long}, though LSTMs do not reach the same level of performance in our setting. Moreover, we counter the scattered annotations available from user-driven associations by training the model largely based on the entire unannotated corpus. The encoder is coupled with a decoder and trained to reproduce the title from the noisy question body. The methodology is reminiscent of recent encoder-decoder networks in machine translation and document summarization~\cite{kalchbrennerB13,sutskever2014sequence,cho2014learning,rush2015neural}. The resulting encoder is subsequently fine-tuned discriminatively on the basis of limited annotations yielding an additional performance boost. 
 


We evaluate our model on the AskUbuntu corpus from Stack Exchange used in prior work~\cite{dossantos-EtAl:2015}. During training, we directly utilize noisy pairs readily available in the forum, but 
to have a realistic evaluation of the system performance, we manually annotate 8K pairs of questions. This clean data is used in two splits, one for development and hyper parameter tuning and another for testing. We evaluate our model and the baselines using standard information retrieval (IR) measures such as Mean Average Precision (MAP), Mean Reciprocal Rank (MRR) and Precision at $n$ (P@$n$). Our full model achieves a \add{MRR of 75.6\% and P@1 of 62.0\%, yielding 8\%  absolute
improvement over a standard IR baseline, and 4\% over standard neural network architectures (including CNNs, LSTMs and GRUs).} 

%

\section{Related Work}

Given the growing popularity of community QA forums, question retrieval has emerged as an important area of research~\cite{nakov-EtAl:2015:SemEval,nakov-EtAl:2016:SemEval}. Previous work on question retrieval has modeled this task using machine translation,  topic modeling and knowledge graph-based approaches ~\cite{jeon2005finding,li2011improving,duan2008searching,zhou2013improving}. More recent work relies on representation learning to go beyond word-based methods. For instance, \newcite{zhou-EtAl:2015} learn word embeddings using category-based metadata information for questions. They define each question as a distribution which generates each word (embedding) independently, and subsequently use a Fisher kernel to assess question similarities. Dos Santos et al.~\shortcite{dossantos-EtAl:2015} propose an approach which combines a convolutional neural network (CNN) and a bag-of-words representation for comparing questions. In contrast to~\cite{zhou-EtAl:2015}, our model treats each question as a word sequence as opposed to a bag of words, and we apply a recurrent convolutional model as opposed to the traditional CNN model used by~\newcite{dossantos-EtAl:2015} to map questions into meaning representations. Further, we propose a training paradigm that utilizes the entire corpus of unannotated questions in a semi-supervised manner.

Recent work on answer selection 
on community QA forums, similar to our task of question retrieval, has also involved the use of neural network architectures~\cite{severyn2015sigir,wang-nyberg:2015:ACL-IJCNLP,shen2015word,feng2015applying,tan2015lstm}.  Compared to our work, these approaches focus on improving various other aspects of the model. For instance, \newcite{feng2015applying} explore different similarity measures beyond
cosine similarity, and \newcite{tan2015lstm} adopt the neural attention mechanism over RNNs to generate better answer representations given the questions as context.

\section{Question Retrieval Setup}
\label{s-formulation}

We begin by introducing the basic discriminative setting for retrieving similar
questions. Let $q$ be a query question which generally consists of both a title
sentence and a body section.  For efficiency reasons, we do not compare $q$
against all the other queries in the data base.  Instead, we retrieve first a
smaller candidate set of related questions $Q(q)$ using a standard IR engine,
and then we apply the more sophisticated models only to this reduced set.  Our
goal is to rank the candidate questions in $Q(q)$ so that all the similar
questions to $q$ are ranked above the dissimilar ones. To do so, we define a
similarity score $s(q,p;\theta)$ with parameters $\theta$, where the similarity
measures how closely candidate $p\in Q(q)$ is related to question $q$. The
method of comparison can make use of the title and body of each question. 

The scoring function $s(\cdot,\cdot;\theta)$ can be optimized on the basis of
annotated data $D=\left\lbrace(q_i, p^{+}_i, Q^-_i)\right\rbrace$, where
$p^{+}_i$ is a question similar to question $q_i$ and $Q^-_i$ is a negative set
of questions deemed not similar to $q_i$.  During training, the correct pairs of
similar questions are obtained from available user-marked pairs, while the
negative set $Q^-_i$ is drawn randomly from the entire corpus with the idea that
the likelihood of a positive match is small given the size of the corpus.  The
candidate set during training is just $Q(q_i)=\{p^+_i\}\cup Q^-_i$.  During
testing, the candidate sets are retrieved by an IR engine and we evaluate
against explicit manual annotations. 

In the purely discriminative setting, we use a max-margin framework for learning (or fine-tuning) parameters $\theta$. Specifically, in a context of a particular training example where $q_i$ is paired with $p^+_i$, we minimize the max-margin loss $L(\theta)$ defined as 
\begin{align*}
\max_{p\in Q(q_i)}\left\lbrace s(q_i, p;\theta) - s(q_i, p^+_i;\theta) +
    \delta(p,p^+_i) \right\rbrace,
\end{align*}
where $\delta(\cdot,\cdot)$ denotes a non-negative margin. We set $\delta(p, p^+_i)$
to be a small constant when $p\neq p^+_i$ and 0 otherwise. The parameters $\theta$ can be optimized through sub-gradients $\partial{L}/\partial{\theta}$ aggregated over small batches of the training instances. 

\begin{figure}[t!]
\centering
\includegraphics[width=2.5in]{\figpath 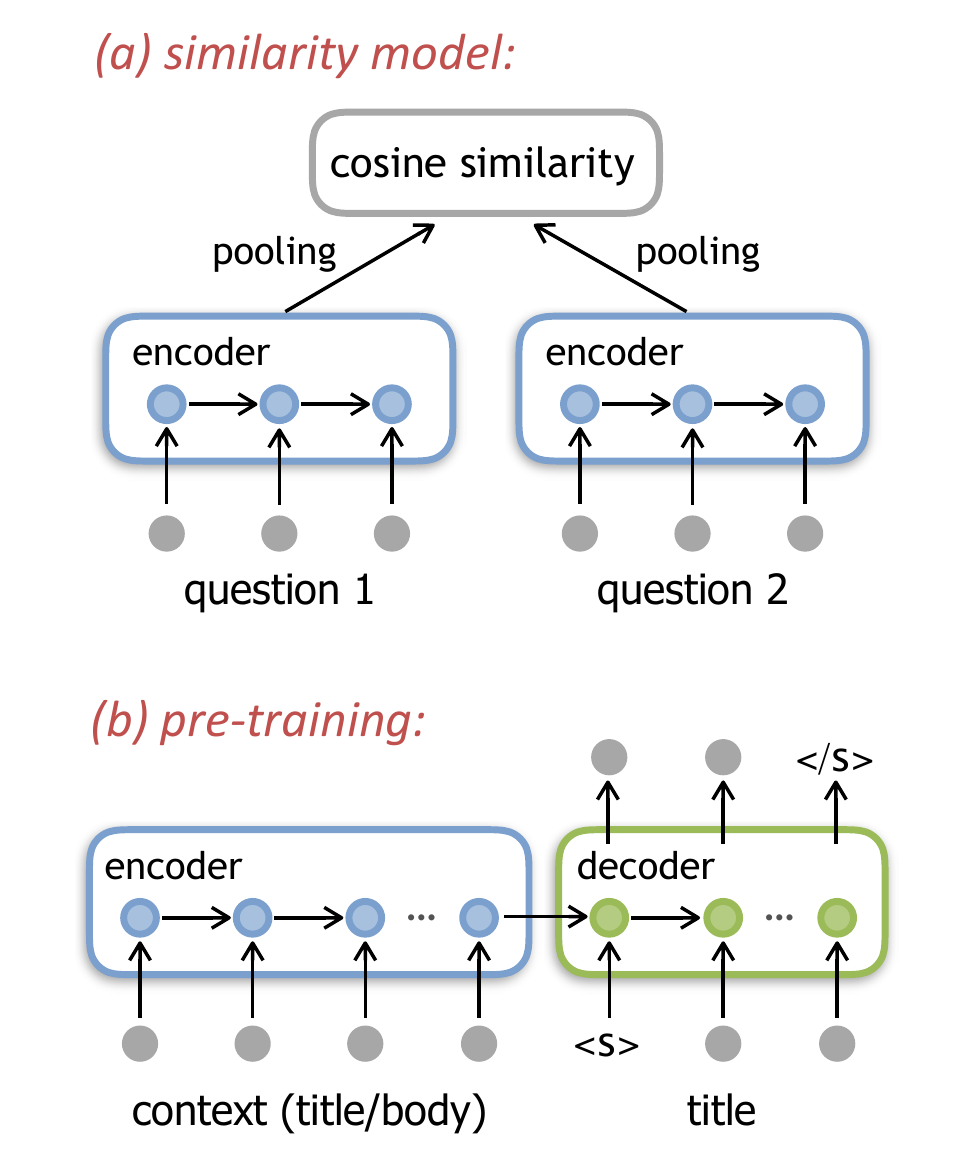}
\vspace{-0.1in}
\caption{Illustration of our model.}
\label{fig:overview}
\end{figure}

There are two key problems that remain. First, we have to define and parameterize the scoring function $s(q,p;\theta)$. We design a recurrent neural network model for this purpose and use it as an \emph{encoder} to map each question into its meaning representation. The resulting similarity function $s(q,p;\theta)$ is just the cosine similarity between the corresponding representations, \add{as shown in Figure~\ref{fig:overview}~(a)}. The parameters $\theta$ pertain to the neural network only. Second, in order to offset the scarcity and limited coverage of the training annotations, we pre-train the parameters $\theta$ on the basis of the much larger unannotated corpus. The resulting parameters are subsequently fine-tuned using the discriminative setup described above.

\section{Model}

\subsection{Non-consecutive Convolution}

We describe here our encoder model, i.e., the method for mapping the question title and body to a vector representation. Our approach is inspired by temporal convolutional neural networks~\cite{lecun-98} and, in particular, its recent refinement~\cite{Lei15}, tailored to capture longer-range, non-consecutive patterns in a weighted manner. Such models can be used to effectively summarize occurrences of patterns in text and aggregate them into a vector representation. However, the summary produced is not selective since all pattern occurrences are counted, weighted by how cohesive (non-consecutive) they are. 
In our problem, the question body tends to be very long and full of irrelevant words and fragments. Thus, we believe that interpreting the question body requires a more selective approach to pattern extraction. 

Our model successively reads tokens in the question title or body, denoted as $\{\mbf{x}_i\}_{i=1}^l$, and transforms this sequence into a sequence of states $\{\mbf{h}_i\}_{i=1}^l$. The resulting state sequence is subsequently aggregated into a single final vector representation for each text as discussed below. Our approach builds on \cite{Lei15}, thus we begin by briefly outlining it. Let $W_1$ and $W_2$ denote filter matrices (as parameters) for pattern size $n=2$. \newcite{Lei15} generate a sequence of states in response to tokens according to

\begin{small}
\vspace{-0.15in}
\begin{align*}
    \mbf{c}_{t',t} &= \mbf{W}_1\mbf{x}_{t'} + \mbf{W}_2\mbf{x}_t \\
    \mbf{c}_t      &= \sum_{t'<t} \lambda^{t-t'-1} \mbf{c}_{t',t} \\
    \mbf{h}_t      &= \tanh(\mbf{c}_t + \mbf{b})
\end{align*}
\end{small}
where $\mbf{c}_{t',t}$ represents a bigram pattern, $\mbf{c}_t$ accumulates a range of patterns and $\lambda\in [0,1)$ is a constant decay factor used to down-weight patterns with longer spans. The operations can be cast in a ``recurrent'' manner and evaluated with dynamic programming. The problem with the approach for our purposes is, however, that the weighting factor $\lambda$ is the same (constant) for all, not triggered by the state $\mbf{h}_{t-1}$ or the observed token $\mbf{x}_t$. 

\paragraph{Adaptive Gated Decay}
We refine this model by learning context dependent weights. For example, if the current input token provides no relevant information (e.g., symbols, functional words), the model should ignore it by incorporating the token with a vanishing weight. In contrast, strong semantic content words such as ``ubuntu'' or ``windows'' should be included with much larger weights. To achieve this effect we introduce \textit{neural gates} similar to LSTMs to specify when and how to average the observed signals. The resulting architecture integrates recurrent networks with non-consecutive convolutional models:

\begin{small}
\vspace{-0.15in}
\begin{align*}
    \mbf{\lambda}_t &= \sigma(\mbf{W}^{\lambda}\mbf{x}_t +
    \mbf{U}^\lambda\mbf{h}_{t-1}+\mbf{b}^\lambda) \\
    \mbf{c}^{(1)}_t &= \mbf{\lambda}_t\odot \mbf{c}^{(1)}_{t-1} +
    (1-\mbf{\lambda}_t)\odot (\mbf{W}_1\mbf{x}_t) \\
    \mbf{c}^{(2)}_t &= \mbf{\lambda}_t\odot \mbf{c}^{(2)}_{t-1} +
    (1-\mbf{\lambda}_t)\odot (\mbf{c}^{(1)}_{t-1}+\mbf{W}_2\mbf{x}_t) \\
    &\cdots \\
    \mbf{c}^{(n)}_t &= \mbf{\lambda}_t\odot \mbf{c}^{(n)}_{t-1} +
    (1-\mbf{\lambda}_t)\odot (\mbf{c}^{(n-1)}_{t-1}+\mbf{W}_n\mbf{x}_t) \\
    \mbf{h}_t &= \tanh(\mbf{c}^{(n)}_t + \mbf{b})
\end{align*}
\end{small}
where $\sigma(\cdot)$ is the sigmoid function and $\odot$ represents the element-wise product. Here $\mbf{c}^{(1)}_t, \cdots, \mbf{c}^{(n)}_t$ are accumulator vectors
that store weighted averages of 1-gram to $n$-gram features. When the gate $\lambda_t =0$ (vector) for
all $t$, the model represents a traditional CNN with filter width $n$. As $\lambda_t >0$, however, $\mbf{c}^{(n)}_t$ becomes the sum of an exponential number of terms, enumerating all possible
$n$-grams within $\mbf{x}_1,\cdots,\mbf{x}_t$ (seen by expanding the formulas). Note that the gate $\lambda_t(\cdot)$ is parametrized and responds directly to the previous state and the token in question. We refer to this model as RCNN from here on.

\paragraph{Pooling}
In order to use the model as part of the discriminative question retrieval framework outlined earlier, we must condense the state sequence to a single vector. There are two simple alternative \emph{pooling} strategies that we have explored -- either averaging over the states\footnote{We also normalize state vectors before
averaging, which empirically gets better performance.} or simply taking the last
one as the meaning representation. In addition, we apply the encoder to both the question title and body, and the final representation is computed as the average of the two resulting vectors. 

Once the aggregation is specified, the parameters of the gate and the filter matrices can be learned in a purely discriminative fashion. Given that the available annotations are limited and user-guided, we instead use the discriminative training only for fine tuning an already trained model. The method of pre-training the model on the basis of the entire corpus of questions is discussed next. 

\subsection{Pre-training Using the Entire Corpus}
The number of questions in the AskUbuntu corpus far exceeds user annotations of pairs of similar questions. We can make use of this larger raw corpus in two different ways. First, since models take word embeddings as input we can tailor the embeddings to the specific vocabulary and expressions in this corpus. To this end, we run word2vec~\cite{mikolov2013} on the raw corpus in addition to the Wikipedia dump. Second, and more importantly, we use individual questions as training
examples for an auto-encoder constructed by pairing the encoder model (RCNN) with an corresponding decoder (of the same type). \add{As illustrated in Figure~\ref{fig:overview}~(b),} the resulting encoder-decoder architecture is akin to those used in machine translation~\cite{kalchbrennerB13,sutskever2014sequence,cho2014learning} and summarization~\cite{rush2015neural}.

Our encoder-decoder pair represents a conditional language model
$P(\text{title}\vert\text{context})$, where the context can be any of (a) the original
title itself, (b) the question body and (c) the title/body of a similar
question. All possible (title, context) pairs are used during training to
optimize the likelihood of the words (and their order) in the titles. We use the question title as the target for two reasons. The question body contains more information than the title but also has many irrelevant details. As a result, we can view the title as a distilled summary of the noisy body, and the encoder-decoder model is trained to act as a de-noising auto-encoder. Moreover, training a decoder for the title (rather than the body) is also much faster since titles tend to be short (around 10 words).

The encoders pre-trained in this manner are subsequently fine-tuned according to the discriminative criterion described already in Section~\ref{s-formulation}.

\section{Alternative models}

For comparison, we also train three alternative benchmark encoders (LSTMs, GRUs and CNNs) for mapping questions to vector representations.
LSTM and GRU-based encoders can be pre-trained analogously to RCNNs, and fine-tuned discriminatively. CNN encoders, on the other hand, are only trained discriminatively. 
While plausible, neither alternative reaches quite the same level of performance as our pre-trained RCNN. 
%
%
\paragraph{LSTMs}
LSTM cells~\cite{hochreiter1997long} have been used to capture
semantic information across a wide range of applications, including
machine translation and entailment
recognition~\cite{bahdanau2014neural,bowman2015large,rocktaschel2015reasoning}.
Their success can be attributed to neural gates that adaptively read or discard information
to/from internal memory states. 

Specifically, a LSTM network successively reads the input token $\mbf{x}_t$,
internal state $\mbf{c}_{t-1}$, as well as the visible state $\mbf{h}_{t-1}$,
and generates the new states $\mbf{c}_t, \mbf{h}_t$:

\begin{small}
\vspace{-0.15in}
\begin{align*}
    \mbf{i}_t &= \sigma(\mbf{W}^i\mbf{x}_t + \mbf{U}^i\mbf{h}_{t-1} + \mbf{b}^i) \\
    \mbf{f}_t &= \sigma(\mbf{W}^f\mbf{x}_t + \mbf{U}^f\mbf{h}_{t-1} + \mbf{b}^f) \\
    \mbf{o}_t &= \sigma(\mbf{W}^o\mbf{x}_t + \mbf{U}^o\mbf{h}_{t-1} + \mbf{b}^o) \\
    \mbf{z}_t &= \tanh(\mbf{W}^z\mbf{x}_t + \mbf{U}^z\mbf{h}_{t-1} + \mbf{b}^z) \\
    \mbf{c}_t &= \mbf{i}_t\odot\mbf{z}_t + \mbf{f}_t\odot\mbf{c}_{t-1} \\
    \mbf{h}_t &= \mbf{o}_t\odot\tanh(\mbf{c}_t)
\end{align*}
\end{small}
where $\mbf{i}$, $\mbf{f}$ and $\mbf{o}$ are \textit{input}, \textit{forget} and
\textit{output} gates, respectively. Given the visible state sequence
$\{\mbf{h}_i\}_{i=1}^l$, we can aggregate it to a single vector exactly as with
RCNNs. The LSTM encoder can be pre-trained (and fine-tuned) in the similar way as
our RCNN model. \add{For instance, \newcite{dai2015semi} recently adopted
pre-training for text classification task.}

\paragraph{GRUs} A GRU is another comparable unit for sequence
modeling~\cite{cho2014properties,chung2014empirical}. Similar to the LSTM unit,
the GRU has two neural gates that control the flow of information:

\begin{small}
\vspace{-0.15in}
\begin{align*}
    \mbf{i}_t &= \sigma(\mbf{W}^i\mbf{x}_t + \mbf{U}^i\mbf{h}_{t-1} + \mbf{b}^i) \\
    \mbf{r}_t &= \sigma(\mbf{W}^r\mbf{x}_t + \mbf{U}^r\mbf{h}_{t-1} + \mbf{b}^r) \\
    \mbf{c}_t &= \tanh(\mbf{W}\mbf{x}_t + \mbf{U}(\mbf{r}_t\odot\mbf{h}_{t-1})+\mbf{b}) \\
    \mbf{h}_t &= \mbf{i}_t\odot\mbf{c}_t + (1-\mbf{i}_t)\odot\mbf{h}_{t-1}
\end{align*}
\end{small}
where $i$ and $r$ are \textit{input} and \textit{reset} gate respectively. Again, the GRUs
can be trained in the same way.

\paragraph{CNNs}
Convolutional neural networks~\cite{lecun-98} have also been successfully applied to
various NLP
tasks~\cite{kalchbrenner2014,Kim14,kim2015character,zhang2015text,gao2014modeling}.
As models, they are different from LSTMs since the temporal 
convolution operation and associated filters
map \textit{local chunks} (windows) of the input into a feature
representation.  Concretely, if we let $n$ denote the filter width, and
$\mbf{W}_{1},\cdots,\mbf{W}_n$ the corresponding filter matrices, then the convolution
operation is applied to each window of $n$ consecutive words as follows:

\begin{small}
\vspace{-0.15in}
\begin{align*}
    \mbf{c}_t &= \mbf{W}_1\mbf{x}_{t-n+1} + \mbf{W}_2\mbf{x}_{t-n+2} + \cdots
              + \mbf{W}_n\mbf{x}_{t} \\
    \mbf{h}_t &= \tanh(\mbf{c}_t + \mbf{b})
\end{align*}
\end{small}
The sets of output state vectors $\{\mbf{h}_t\}$ produced in this case are
typically referred to as feature maps. Since each vector in the feature map only
pertains to local information, the last vector is not sufficient to capture the
meaning of the entire sequence. Instead, we consider \emph{max-pooling} or
\emph{average-pooling} to obtain the aggregate representation for the entire
sequence.

\section{Experimental Setup}

\paragraph{Dataset} We use the Stack Exchange AskUbuntu dataset used in prior
work~\cite{dossantos-EtAl:2015}. This dataset contains 167,765 unique questions,
each consisting of a title and a body\footnote{\add{We truncate the body section
at a maximum of 100 words.}}, and a set of user-marked similar question pairs.
We provide various statistics from this dataset in Table~\ref{table:stats}.

\begin{table}[t]
\footnotesize
\centering
\begin{tabular}{|c|l|r|}
\hline
\multirow{1}{*}{Corpus}
    & \# of unique questions & 167,765 \\
    & Avg length of title & \add{6.7} \\
    & Avg length of body& \add{59.7} \\
\hline
\multirow{2}{*}{Training} 
    & \# of unique questions & 12,584 \\
    & \# of user-marked pairs & 16,391 \\
\hline
\multirow{3}{*}{Dev}
    & \# of query questions & 200 \\
    & \# of annotated pairs & 200$\times$20 \\
    & Avg \# of positive pairs per query & 5.8 \\
\hline
\multirow{3}{*}{Test}
    & \# of query questions & 200 \\
    & \# of annotated pairs & 200$\times$20 \\
    & Avg \# of positive pairs per query & 5.5 \\
\hline
\end{tabular}
\caption{Various statistics 
from our Training, Dev, and Test sets derived from the Sept. 2014 Stack Exchange AskUbuntu dataset.}
\label{table:stats}
\end{table}

\begin{table*}[!th!]
\small
\centering
\begin{tabular}{l|l|c|c|c|c|c|c|c|c}
\hline
\multirow{2}{*}{Method} & \multirow{2}{*}{Pooling} & \multicolumn{4}{c|}{Dev}  & \multicolumn{4}{c}{Test} \\ \cline{3-10}
		& & MAP  & MRR  & P@1  & P@5  & MAP  & MRR  & P@1  & P@5  \\ \hline
BM25	& - & 52.0 & 66.0 &  51.9 & 42.1 & 56.0 & 68.0 & 53.8 & 42.5 \\
TF-IDF	& - & 54.1 & 68.2 & 55.6 & 45.1 & 53.2 & 67.1 & 53.8 & 39.7 \\
SVM		& - & 53.5 & 66.1 & 50.8 & 43.8 & 57.7 & 71.3 & 57.0 &  43.3    \\
\hline
CNNs & mean & 58.5 & 71.1 & 58.4 & 46.4 & 57.6 & 71.4 & 57.6 & 43.2 \\
LSTMs & mean & 58.4 & 72.3 & 60.0 & 46.4 & 56.8 & 70.1 & 55.8 & 43.2 \\
GRUs   & mean & 59.1 & 74.0 & 62.6 & 47.3 & 57.1 & 71.4 & 57.3 & 43.6 \\
RCNNs  & last & 59.9 & 74.2 & 63.2 & 48.0 & 60.7 & 72.9 & 59.1 & 45.0 \\
\hline
LSTMs + pre-train & mean & 58.3 & 71.5 & 59.3 & 47.4 & 55.5 & 67.0 & 51.1 & 43.4 \\
GRUs + pre-train & last & 59.3 & 72.2 & 59.8 & 48.3 & 59.3 & 71.3 & 57.2 & 44.3 \\
RCNNs + pre-train & last & {~}{\bf 61.3}$^*$ & {\bf 75.2} & {\bf 64.2} & {~}{\bf 50.3}$^*$ & {~}{\bf 62.3}$^*$ & {~}{\bf 75.6}$^*$ & {\bf 62.0} & {~}{\bf 47.1}$^*$ \\
\hline
\end{tabular}
\vspace{0.02in}
\caption{Comparative results of all methods on the question similarity task. \add{Higher numbers are better. For neural network models, we show the best average performance across 5 independent runs and the corresponding pooling strategy. Statistical significance with $p<0.05$ against other types of model is marked with $*$.}} 
\label{table:overall}
\end{table*}


\paragraph{Gold Standard for Evaluation} User-marked similar question pairs on QA sites are often known to be incomplete. In order to evaluate this in our dataset, we took a sample set of questions paired with 20 candidate questions retrieved by a search engine trained on the AskUbuntu data.
The search engine used is the well-known BM25 model \cite{robertson2009probabilistic}. Our manual evaluation of the candidates showed that only 5$\%$ of the similar questions were marked by users, with a precision of 79$\%$. 
%
Clearly, this low recall would not lead to a realistic evaluation if we used
user marks as our gold standard. Instead, we make use of expert annotations
carried out on a subset of questions.



\paragraph{Training Set} We use user-marked similar pairs as positive pairs in training since the annotations have high precision and do not require additional manual annotations. This allows us to use a much larger training set. We use random questions from the corpus paired with each query question $p_i$ as negative pairs in training.  We randomly sample 20 questions as negative examples for each $p_i$ during each epoch.

\paragraph{Development and Test Sets} We re-constructed the new dev and test sets
consisting of the first 200 questions from the dev and test sets provided
by~\newcite{dossantos-EtAl:2015}.  For each of the above questions, we retrieved
the top 20 similar candidates using BM25 and manually
annotated the resulting 8K pairs as similar or non-similar.\footnote{The
    annotation task was initially carried out by two expert annotators,
    independently. The initial set was refined by comparing the annotations and
    asking a third judge to make a final decision on disagreements. After a
    consensus on the annotation guidelines was reached (producing a Cohen's
    kappa 
    of 0.73), the overall annotation was carried out by only one expert.}


\paragraph{Baselines and Evaluation Metrics}
We evaluated neural network models---including \textbf{CNNs}, \textbf{LSTMs},
\textbf{GRUs} and \textbf{RCNNs}---by comparing them with the following
baselines:
\begin{itemize}
\vspace{-0.11in}
    \item{\bf BM25}, we used the BM25 similarity measure provided by Apache Lucene.
\vspace{-0.11in}
    \item{\bf TF-IDF}, we ranked questions using cosine similarity based on a vector-based word representation for each question.
\vspace{-0.11in}
    \item{\bf SVM}, we trained a re-ranker using SVM-Light \cite{Joachims02c} with a linear kernel incorporating several similarity measures from the DKPro similarity package~\cite{bar-zesch-gurevych:2013:SystemDemo}. 
 \end{itemize}
\vspace{-0.11in}
We evaluated the models based on the following IR metrics: Mean Average Precision (MAP), Mean Reciprocal Rank (MRR), Precision at 1 (P@1), and Precision at 5 (P@5). 

\begin{table}[t]
\small
\centering
\begin{tabular}{lccc}
\hline
 & $d$ & $\vert\theta\vert$ & $n$ \\
\hline
LSTMs & 240 & 423K & - \\
GRUs & 280 & 404K & - \\
CNNs & 667 & 401K & 3 \\
RCNNs & 400 & 401K & 2 \\
\hline
\end{tabular}
\caption{Configuration of neural models. $d$ is
the hidden dimension, $\vert\theta\vert$ is the number of parameters and $n$ is
the filter width.}
\label{table:config}
\end{table}

\paragraph{Hyper-parameters}
We performed an extensive hyper-parameter search to identify the best model for
the baselines and neural network models. For the \textbf{TF-IDF} baseline, we
tried $n$-gram feature order $n\in \{1,2,3\}$ with and without stop words
pruning.  For the \textbf{SVM} baseline, we used the default SVM-Light
parameters whereas the dev data is only used to increase the training set size
when testing on the test set. We also tried to give higher weight to dev
instances but this did not result in any improvement.

For all the neural network models, we used Adam~\cite{kingma2015adam} as the
optimization method with the default setting suggested by the authors. We
optimized other hyper-parameters with the following range of values: learning
rate $\in \{ 1e-3, 3e-4 \}$, dropout~\cite{dropout} probability $\in \{ 0.1,
0.2, 0.3 \}$, CNN feature width $\in \{2, 3, 4\}$. We also tuned the pooling
strategies and ensured each model has a comparable number of parameters. The
default configurations of LSTMs, GRUs, CNNs and RCNNs are shown in
Table~\ref{table:config}. We used MRR to identify the best training epoch and
the model configuration. For the same model configuration, we report average
performance across 5 independent runs.\footnote{\add{For a fair comparison, we
also pre-train 5 independent models for each configuration and then fine tune
these models. We use the same learning rate and dropout rate during pre-training
and fine-tuning.}}

\paragraph{Word Vectors} We ran word2vec~\cite{mikolov2013} to obtain
200-dimensional word embeddings using all Stack Exchange data (excluding
StackOverflow) and a large Wikipedia corpus. The word vectors are fixed to avoid
over-fitting across all experiments.

\section{Results}

\begin{table*}[!t!h!]
\small
\vspace{-0.1in}
\centering
\begin{tabular}{l|c|c|c|c|c|c|c|c}
\hline
\multirow{2}{*}{Method} & \multicolumn{4}{c|}{Dev}  & \multicolumn{4}{c}{Test} \\ \cline{2-9}
 & MAP  & MRR  & P@1  & P@5  & MAP  & MRR  & P@1  & P@5  \\ \hline
CNNs, max-pooling & 57.8 & 69.9 & 56.6 & 47.7 & 59.6 & 73.1 & 59.6 & 45.4 \\
CNNs, mean-pooling & 58.5 & 71.1 & 58.4 & 46.4 & 57.6 & 71.4 & 57.6 & 43.2 \\
\hline
LSTMs + pre-train, mean-pooling & 58.3 & 71.5 & 59.3 & 47.4 & 55.5 & 67.0 & 51.1 & 43.4 \\
LSTMs + pre-train, last state & 57.6 & 71.0 & 58.1 & 47.3 & 57.6 & 69.8 & 55.2 & 43.7 \\
\hline
GRUs + pre-train, mean-pooling & 57.5 & 69.9 & 57.1 & 46.2 & 55.5 & 67.3 & 52.4 & 42.8 \\
GRUs + pre-train, last state & 59.3 & 72.2 & 59.8 & 48.3 & 59.3 & 71.3 & 57.2 & 44.3 \\
\hline
RCNNs + pre-train, mean-pooling & 59.3 & 73.6 & 61.7 & 48.6 & 58.9 & 72.3 & 57.3 & 45.3 \\
RCNNs + pre-train, last state & {\bf 61.3} & {\bf 75.2} & {\bf 64.2} & {\bf 50.3} & {\bf 62.3} & {\bf 75.6} & {\bf 62.0} & {\bf 47.1} \\
\hline
\end{tabular}
\caption{Choice of pooling strategies.}
\label{table:pooling}
\end{table*}

\paragraph{Overall Performance}
Table~\ref{table:overall} shows the performance of the baselines and the neural
encoder models on the question retrieval task.  The results show that our full
model, RCNNs with pre-training, achieves the best performance across all metrics
on both the dev and test sets. For instance, the full model gets a P@1 of 62.0\%
on the test set, outperforming the word matching-based method BM25 by over 8 
percent points. 
Further, our RCNN model also outperforms the other neural encoder models and the
baselines across all metrics.  This superior performance indicates that the use
of non-consecutive filters and a varying decay is effective in improving
traditional neural network models.

Table~\ref{table:overall} also demonstrates the performance gain from
pre-training the RCNN encoder. The RCNN model when pre-trained on the entire
corpus consistently gets better results across all the metrics.


\paragraph{Pooling Strategy} We analyze the effect of various pooling strategies
for the neural network encoders. As shown in Table~\ref{table:pooling}, our RCNN
model outperforms other neural models regardless of the two pooling strategies
explored. We also observe that simply using the last hidden state as the
final representation achieves better results for the RCNN model.

\paragraph{Using Question Body} Table~\ref{table:body} compares the performance of the
TF-IDF baseline and the RCNN model when using question titles only or when using
question titles along with question bodies. TF-IDF's performance
changes very little when the question bodies are included (MRR and P@1 are slightly 
better but MAP is slightly worse).
However, we find that the inclusion of the question bodies improves the
performance of the RCNN model, achieving a 1\% to 3\% improvement with both
model variations. The RCNN model's greater improvement illustrates the ability
of the model to pick out components that pertain most directly to the question
being asked from the long, descriptive question bodies.

\begin{table}[!t]
\footnotesize
\centering
\begin{tabular}{l|c|c|c}
\hline
TF-IDF & MAP  & MRR  & P@1  \\
\hline
$\quad$ title only & 54.3 & 66.8 & 52.7 \\
$\quad$ title + body & 53.2 & 67.1 & 53.8 \\
\hline
\hline
RCNNs, mean-pooling & MAP  & MRR  & P@1 \\
\hline
$\quad$ title only & 56.0 & 68.9 & 55.7 \\
$\quad$ title + body & 58.5 & 71.7 & 56.7\\
\hline
\hline
RCNNs, last state & MAP  & MRR  & P@1 \\
\hline
$\quad$ title only & 58.2 & 70.7 & 56.6 \\
$\quad$ title + body & 60.7 & 72.9 & 59.1 \\
\hline
\end{tabular}
\caption{Comparision between model variants on the test set when question bodies
are used or not used.}
\label{table:body}
\end{table}

\begin{figure}[!t!]
\centering
\includegraphics[width=2.4in]{\figpath 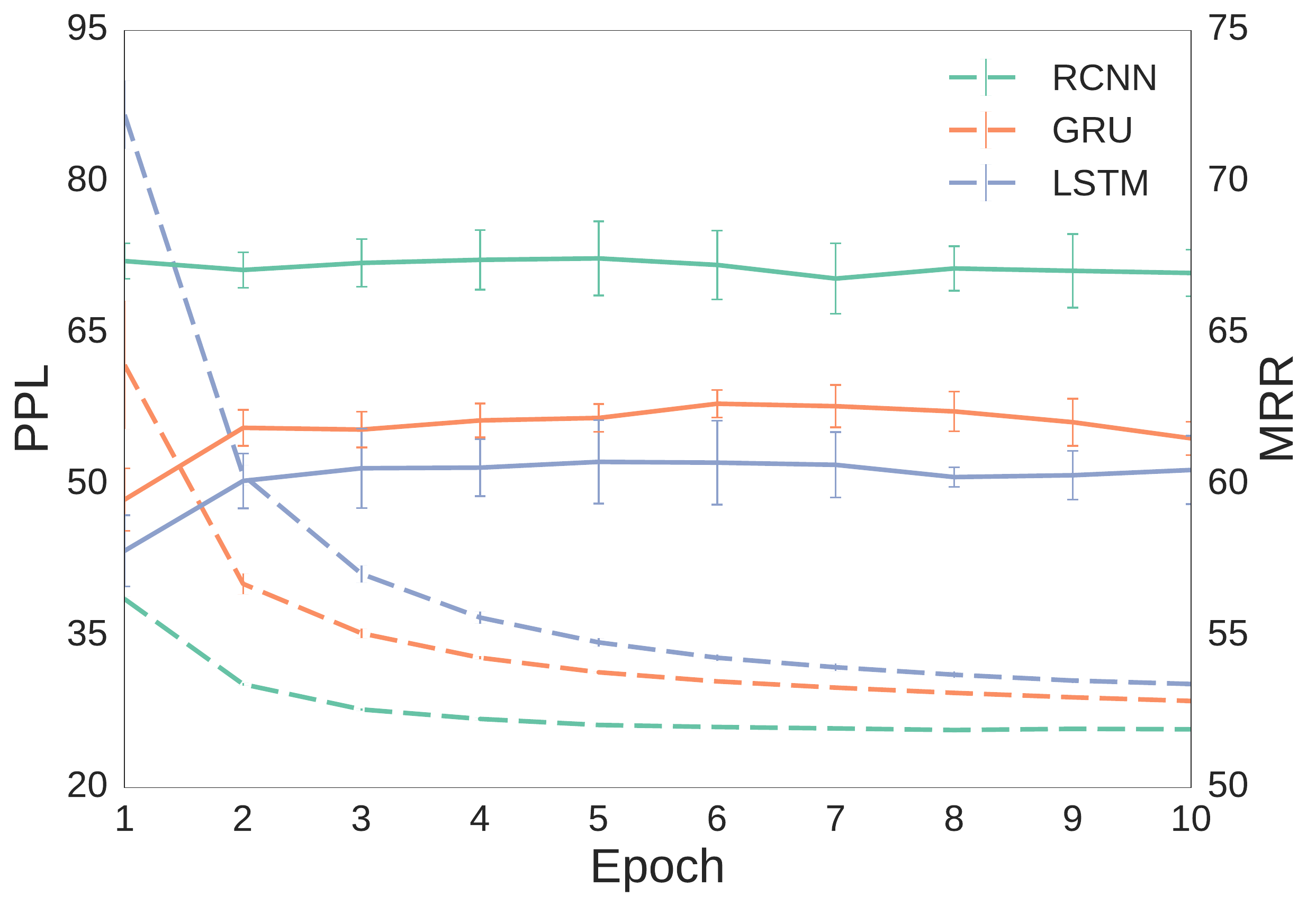}
\vspace{-0.12in}
\caption{\add{Perplexity (dotted lines) on a heldout portion of the corpus versus MRR on the dev set (solid lines) during pre-training. Variances across 5 runs are shown as vertical bars.}}
\label{fig:pretrain}
\end{figure}

\begin{figure*}[!t!h!]
\small
\centering
\begin{tabular}{cc}
\includegraphics[height=0.90in]{\figpath 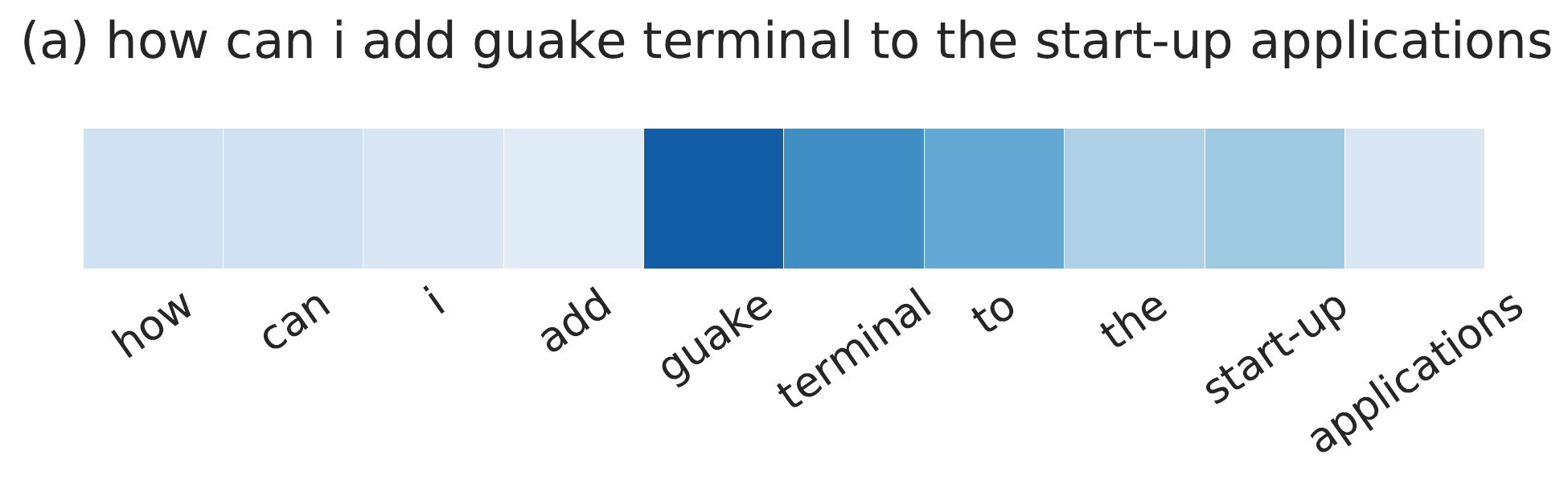} &
\includegraphics[height=0.90in]{\figpath 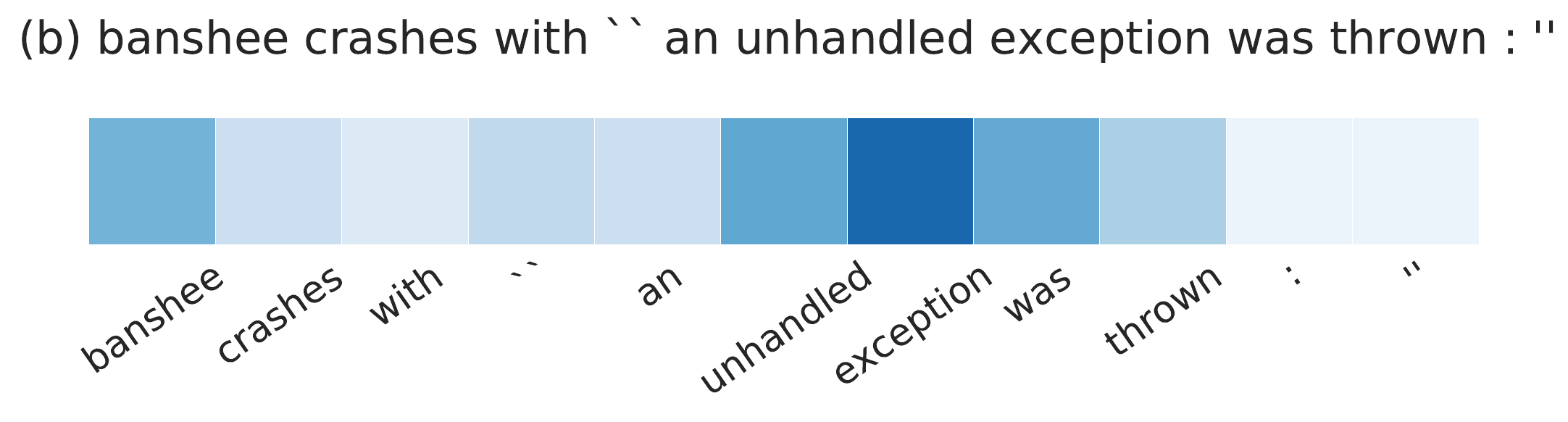} \\
\multicolumn{2}{c}{\includegraphics[height=0.86in]{\figpath 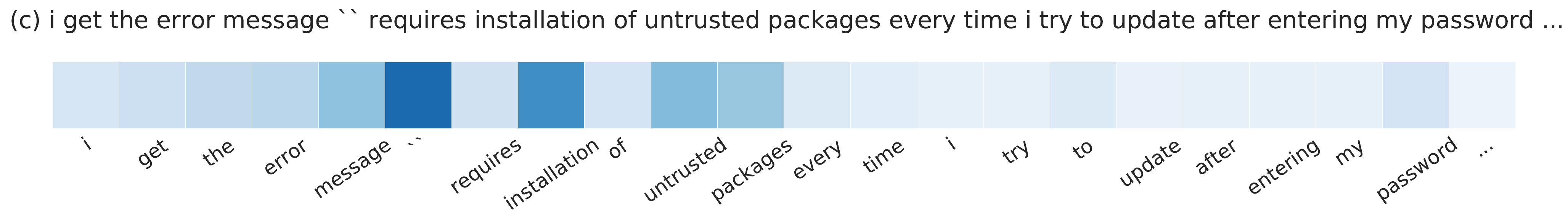}} \\
\multicolumn{2}{c}{\includegraphics[height=0.86in]{\figpath 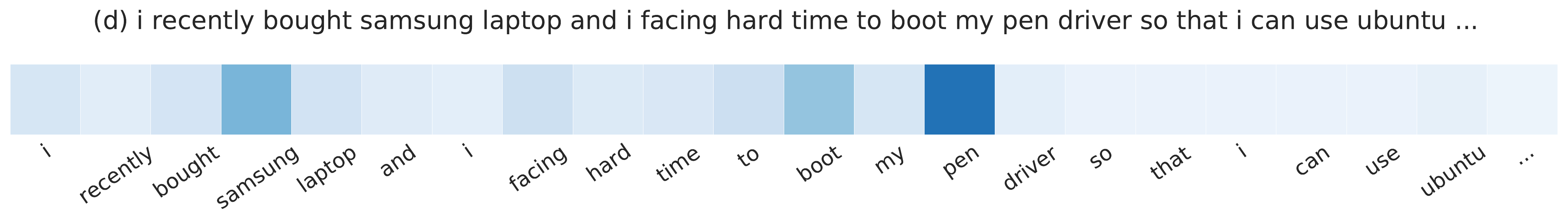}}
\end{tabular}
\vspace{-0.15in}
\caption{Visualizations of $1-\lambda_t$ of our model on several question pieces from the data set. $\lambda_t$ is set to a \emph{scalar value} (instead of 400-dimension vector) to make the visualization simple. The corresponding model is a simplified variant, which is about 4\% worse than our full model.}
\label{fig:casestudy}
\end{figure*}

\paragraph{Pre-training} 
Note that, during pre-training, the last hidden states generated by the neural
encoder are used by the decoder to reproduce the question titles. It would be
interesting to see how such states capture the meaning of questions. \add{To
this end, we evaluate MRR on the dev set using the last hidden states of the
question titles. We also test how the encoder captures information from the
question bodies to produce the distilled summary, i.e. titles. To do
so, we evaluate the perplexity of the trained encoder-decoder model on a heldout
set of the corpus, which contains about 2000 questions.} 

\add{As shown in Figure~\ref{fig:pretrain}, the representations generated by the
RCNN encoder perform quite well, resulting in a perplexity of 25 and over 68\%
MRR without the subsequent fine-tuning. Interestingly, the LSTM and GRU networks
obtain similar perplexity on the heldout set, but achieve much worse MRR for
similar question retrieval. For instance, the GRU encoder obtains only 63\% MRR,
5\% worse than the RCNN model's MRR performance. As a result, the LSTM and GRU
encoder do not benefit clearly from pre-training, as suggested in
Table~\ref{table:overall}.}

\add{The inconsistent performance difference may be explained by two hypotheses.
One is that the perplexity is not suitable for measuring the similarity of the
encoded text, thus the power of the encoder is not illustrated in terms of
perplexity.  Another hypothesis is that the LSTM and GRU encoder may learn
non-linear representations therefore their semantic relatedness can not be
directly accessed by cosine similarity. }

\begin{figure}[!t!]
\centering
\includegraphics[width=2.9in]{\figpath 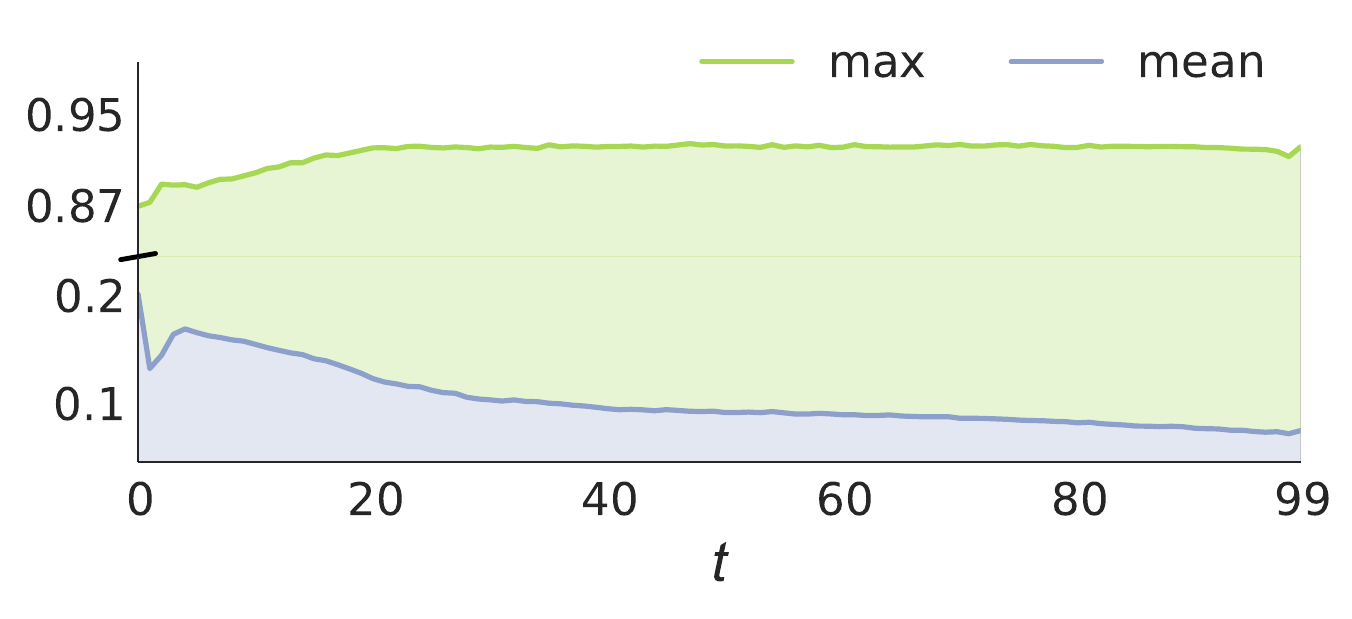}
\vspace{-0.15in}
\caption{\add{The maximum and mean value of the 400-dimentional weight vector
$1-\lambda_t$ at each step (word position) $t$. Values are averaged across all
questions in the dev and test set.}}
\label{fig:decay}
\end{figure}

\paragraph{Adaptive Decay}
\add{Finally, we analyze the gated convolution of our model.
Figure~\ref{fig:decay} demonstrates at each word position $t$ how much input
information is taken into the model by the adaptive weights $1-\lambda_t$. The
average of weights in the vector decreases as $t$ increments, suggesting that
the information encoded into the state vector saturates when more input are
processed. On the other hand, the largest value in the weight vector remains
high throughout the input, indicating that at least some information has been
stored in $\mbf{h}_t$ and $\mbf{c}_t$.}

\add{We also conduct a case study on analyzing the neural gate. Since directly
inspecting the 400-dimensional decay vector is difficult, we train a model that
uses a \emph{scalar decay} instead. As shown in Figure~\ref{fig:casestudy}, the
model learns to assign higher weights to application names and quoted error
messages, which intuitively are important pieces of a question in the AskUbuntu
domain.}

\section{Conclusion}
In this paper, we employ gated (non-consecutive) convolutions to map questions to their semantic representations, and demonstrate their effectiveness on the task of question retrieval in community QA forums. This architecture enables the model to glean
key pieces of information from lengthy, detail-riddled user questions.  Pre-training within an encoder-decoder framework (from body to title) on the basis of the entire raw corpus is integral to the model's success.

\section*{Acknowledgments}
\add{We thank Yu Zhang, Yoon Kim, Danqi Chen, the MIT NLP group
and the reviewers for their helpful comments. The work is developed in
collaboration with the Arabic Language Technologies (ALT) group at Qatar
Computing Research Institute (QCRI) within the {\sc Iyas} project. Any opinions, findings, conclusions, or
recommendations expressed in this paper are those of the authors, and do not
necessarily reflect the views of the funding organizations.}

\bibliography{paper}
\bibliographystyle{naaclhlt2016}

\onecolumn{
\large

\section*{Note on Data Pre-processing Issue:} 

This paper is based on our early preprint version "Denoising Bodies to Titles: Retrieving Similar Questions with Recurrent Convolutional Models". 

$ $

We noticed, during the analysis of the models, that some question text in the AskUbuntu dataset contain user-annotated information which should be removed for a \emph{more realistic} evaluation (see the figure below and the explanation). 

$ $

We re-processed the dataset and updated the results accordingly. The performance of various neural network models reported in this version may not match the numbers reported in our early version exactly, but the relative observation (i.e. relative difference between models) does not change.

$ $ 

$ $

\begin{figure*}[hb]
\centering
\includegraphics[width=5in]{\figpath 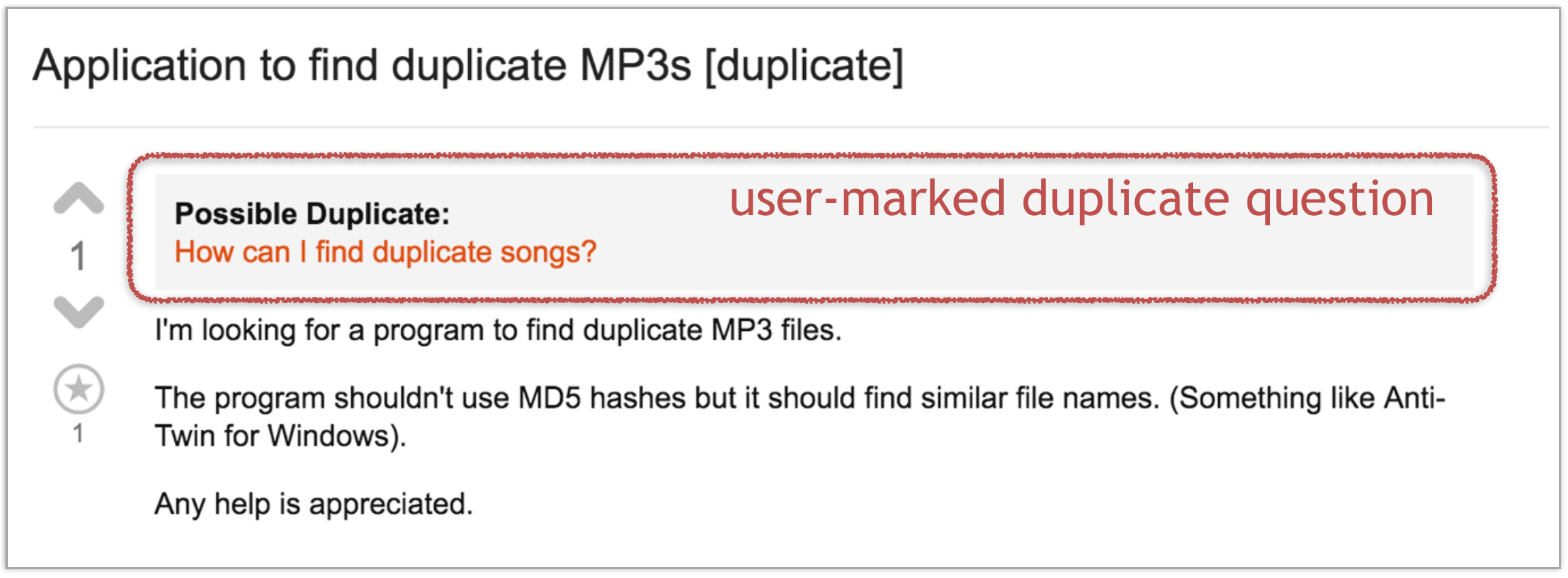}
\captionsetup{labelformat=empty}
\caption*{\normalsize A snapshot of one AskUbuntu question. The "possible duplicate" section shows the title of a similar question marked by the forum user. This section is included in the original XML dump of AskUbuntu, and was not removed in our early experiments.}
\end{figure*}

}

\end{document}